\documentclass[letterpaper, 10pt, conference]{ieeeconf}
\IEEEoverridecommandlockouts

\usepackage{times} 
\usepackage{stfloats}
\usepackage{amsmath} 
\usepackage{amssymb} 
\usepackage[final]{changes} 
\usepackage{graphicx}
\usepackage{widetext}
\usepackage{afterpage}
\usepackage{algorithm}
\usepackage[algo2e, ruled,vlined,noend]{algorithm2e}

\let\labelindent\relax
\usepackage{enumitem}
\usepackage[acronym]{glossaries}
\glsdisablehyper
\usepackage{xcolor}

\usepackage{verbatim}

\usepackage[normalem]{ulem}

\renewcommand{\baselinestretch}{0.97}

\usepackage[natbib=true,backend=bibtex,uniquelist=false,maxnames=2,maxcitenames=2,citestyle=numeric-comp, sorting=none, giveninits=true]{biblatex} \DeclareFieldFormat{apacase}{#1} 
\title{Learning and Planning for Temporally Extended Tasks in Unknown Environments}

\newcommand{\setor}{~|~}
\newcommand{\hist}{h}

\newacronym{dfa}{DFA}{Deterministic Finite Automaton}
\newacronym{sccs}{SCCs}{Strongly Connected Components}
\newacronym{mdp}{MDP}{Markov Decision Process}
\newacronym{fts}{FTS}{Finite Transition System}
\newacronym{ltl}{LTL}{Linear Temporal Logic}
\newacronym{pctl}{PCTL}{Probabilistic Computation Tree Logic}
\newacronym{wdfa}{WDFA}{Weighted Deterministic Finite Automaton}
\newacronym{scc}{SCC}{Strongly Connected Component}
\newacronym{scltl}{scLTL}{Syntactically Co-Safe Linear Temporal Logic}
\newacronym{mcts}{MCTS}{Monte-Carlo Tree-Search}
\newacronym{pouct}{PO-UCT}{Partially Observable UCT}
\newacronym{wfdts}{WFDTS}{Weighted Finite Deterministic Transition System}
\newacronym{lotap}{LoTAP}{Learning over Temporal Actions Planner}
\newacronym{potlp}{PO-TLP}{Partially Observable Temporal Logic Planner}
\newacronym{pomdp}{POMDP}{Partially Observable Markov Decision Process}
\newacronym{lomdp}{LOMDP}{Locally Observable Markov Decision Process}
\newacronym{lts}{LTS}{Labeled Transition System}
\newacronym{lsp}{LSP}{Learned Subgoal Planning}
\newacronym{pa}{PA}{Product Automaton}

\definecolor{purple}{rgb}{0.5,0,0.5}
\definecolor{softblue}{rgb}{0.70,0.70,0.70}

\newcommand{\adam}[1]{{\textcolor{purple}{[Adam] \emph{\bf#1 }}}}
\newcommand{\chris}[1]{{\textcolor{yellow}{[Chris] \emph{\bf#1 }}}}

\newcommand{\sebastian}[1]{{\textcolor{magenta}{[Sebastian] \emph{\bf#1 }}}}

\renewcommand{\adam}[1]{}
\renewcommand{\chris}[1]{}
\renewcommand{\sebastian}[1]{}

\newsavebox\MBox
\newcommand\mathunderline[2][red]{{\sbox\MBox{$#2$}%
  \rlap{\usebox\MBox}\color{#1}\rule[-1.8\dp\MBox]{\wd\MBox}{1.5pt}}}

\newenvironment{flushenumerate}{%
\begin{enumerate}
   {\setlength{\leftmargin}{15pt}}%
    \setlength{\labelwidth}{20pt}
    \setlength{\itemindent}{0pt}
    \setlength{\labelsep}{0.5em}
 \setlength{\itemsep}{1pt}
 \setlength{\parskip}{0pt}
 \setlength{\parsep}{0pt}}
 {\end{enumerate}}

\author{
  Christopher Bradley$^{1}$, Adam Pacheck$^{2}$, Gregory J. Stein$^{3}$, Sebastian Castro$^{1}$, \\
  Hadas Kress-Gazit$^{2}$, and Nicholas Roy$^{1}$
  
\thanks{This work is supported by ONR PERISCOPE MURI  N00014-17-1-2699 and the Toyota Research Institute (TRI).
$^{1}$ CSAIL, Massachusetts Institute of Technology (MIT), Cambridge, MA 02142, USA (\{cbrad, sebacf, nickroy\}@mit.edu). 
$^{2}$ Sibley School of Mechanical and Aerospace Engineering, Cornell University, Ithaca, NY 14850, USA (\{akp84, hadaskg\}@cornell.edu). 
$^{3}$ Computer Science Department, George Mason University, Fairfax, VA 22030, USA (gjstein@gmu.edu).
© 2021 IEEE.  Personal use of this material is permitted.  Permission from IEEE must be obtained for all other uses, in any current or future media, including reprinting/republishing this material for advertising or promotional purposes, creating new collective works, for resale or redistribution to servers or lists, or reuse of any copyrighted component of this work in other works.} 

}

\begin{document}

\maketitle

\begin{abstract}
    
    We propose a novel planning technique for satisfying tasks specified in temporal logic in partially revealed environments.
    We define \emph{high-level actions} derived from the environment and the given task itself, and estimate how each action contributes to progress towards completing the task.
    As the map is revealed, we estimate the cost and probability of success of each action from images and an encoding of that action using a trained neural network. 
    These estimates guide search for \added{the minimum-expected-cost plan within our model}.
    Our learned model is structured to generalize across \added{environments and} task specifications without requiring retraining. 
    We demonstrate an improvement in total cost in both simulated and real-world experiments compared to a heuristic-driven baseline.
\end{abstract}



\section{Introduction}

\begin{figure*}[t]
    \vspace{15pt}
    \centering
    \includegraphics[width=1.0\textwidth]{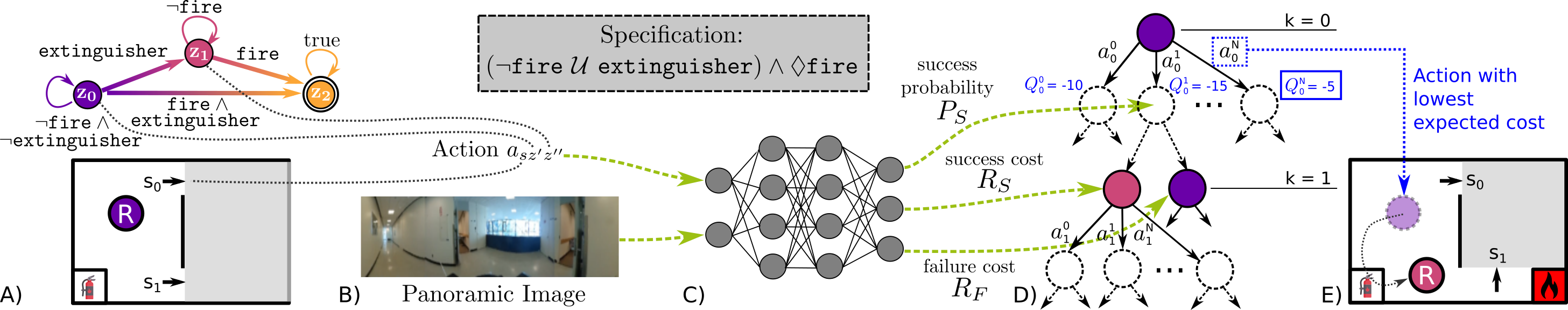}
    \vspace{-18pt}
    \caption{\textbf{Overview.}
    (A) Given a task (e.g., $(\lnot \texttt{fire}\ \mathcal{U}\  \texttt{extinguisher}) \wedge \lozenge \texttt{fire}$), we compute a \gls{dfa} using \cite{spot} to represent the high-level steps needed to accomplish the task.
    The robot operates in a partially explored environment with \textit{subgoals} between observed (white) and unexplored (gray) space, and regions labeled with propositions relevant to the task, e.g., extinguisher and fire.
    (B) We define \textit{high-level actions} with a subgoal (e.g., $s_0$), a \gls{dfa} state $z'$ (e.g., $z_0$) which the robot must be in when arriving at the subgoal, and a state $z''$ (e.g., $z_1$) which the agent attempts to transition to in unknown space.
    (C) For each possible action, we estimate its probability of success ($P_S$) and costs of success ($R_S$) and failure ($R_F$) using a neural network that accepts a panoramic image centered on a subgoal, the distance to that subgoal, and an encoding of the transition from $z'$ to $z''$ (Sec. \ref{sub:encoding}). 
    (D) We estimate the expected cost of each action with these estimates of $P_S$, $R_S$, and $R_F$ using PO-UCT search.
    (E) The agent selects an action with the lowest expected cost, and moves along the path defined by that action, meanwhile receiving new observations, updating its map, and re-planning.
    }
    \label{fig:overview}
    \vspace{-10pt}
\end{figure*}

Our goal is to enable an autonomous agent to find a minimum cost plan for multi-stage planning tasks when the agent's knowledge of the environment is incomplete, i.e., when there are parts of the world the robot has yet to observe.
Imagine, for example, a robot tasked with extinguishing a small fire in a building. To do so, the agent could either find an alarm to trigger the building's sprinkler system, or locate a fire extinguisher, navigate to the fire, and put it out. 
Temporal logic is capable of specifying such complex tasks, and has been used for planning in fully known environments (e.g., \citep{Kress-Gazit2018, Fainekos2009, Bhatia2010, Lacerda2014, Smith2011}).
However, when the environment is initially unknown to the agent, efficiently planning to minimize expected cost to solve these types of tasks can be difficult.

Planning in partially revealed environments poses challenges in reasoning about unobserved regions of space.
Consider our firefighting robot in a building it has never seen before, equipped with perfect perception of what is in direct line of sight from its current position. 
Even with this perfect local sensing, the locations of any fires, extinguishers, and alarms may not be known.
Therefore, the agent must envision all possible configurations of unknown space---including the position of obstacles---to find a plan that satisfies the task specification and, ideally, minimizes cost.
We can model planning under uncertainty in this case as a \gls{lomdp}~\citep{merlin2020locally}, a special class of \gls{pomdp}, which by definition includes our assumption of perfect range-limited perception.
However, planning with such a model requires a distribution over possible environment configurations and high computational effort \citep{Kaelbling:1998:PAP:1643275.1643301,Littman1995362,madani1999pomdp}.


Since finding optimal policies for large \gls{pomdp}s is intractable in general \added{\cite{madani1999pomdp}}, \added{planning to satisfy temporal logic specifications in full \gls{pomdp}s has so far been limited to relatively small environments or time horizons \cite{bouton2020point, ahmadi2020stochastic}.} Thus, many approaches for solving tasks specified with temporal logic often make simplifying assumptions about known and unknown space, \added{or focus on maximizing the  probability a specification is never violated} \cite{Ayala2013temporal, guo2013revising, guo2015multi, Lahijanian2016iterative, Sarid2012guaranteeing}.
Accordingly, such strategies can result in sub-optimal plans, as they do not consider the cost of planning in unknown parts of the map. 


A number of methods use learned policies to minimize the cost of completing tasks specified using temporal logic \citep{icarte2018teaching, sadigh2014learning, Fu2015probably, littman2017ltl, paxton2017combining}. 
However, due to the complexity of these tasks, many are limited to fully observable small grid worlds \citep{icarte2018teaching, sadigh2014learning, Fu2015probably, littman2017ltl} or short time-horizons \citep{paxton2017combining}. 
Moreover, for many of these approaches, changing the specification or environment requires retraining the system. 
Recent work by \citet{Stein2018} uses supervised learning to predict the outcome of acting through unknown space, but is restricted to goal-directed navigation.

\glsreset{scltl}

To address the challenges of planning over a distribution of possible futures, we introduce \gls{potlp}. 
\gls{potlp} enables real-time planning for tasks specified in \gls{scltl} \citep{Kupferman2001} in unexplored, arbitrarily large environments, using an abstraction based on a given task and partially observed map.
Since completing a task may not be possible in known space, we define a set of dynamic \emph{high-level actions} based on transitions between states in our abstraction and available \emph{subgoals} in the map---points on the boundaries between free and unknown space.
We then approximate the full \gls{lomdp} model such that actions either successfully make their desired transitions or fail, splitting future beliefs into two classes and simplifying planning.
We train a neural network from images to predict the cost and outcome of each action, which we use to guide a variant of Monte-Carlo Tree Search (PO-UCT \citep{POMCP}) to find the best high-level action for a given task and observation.

Our model learns directly from visual input and can generalize across novel environments and tasks without retraining.
Furthermore, since our agent continually replans as more of the environment is revealed, we ensure both that the specification is never violated and that we find the optimal trajectory if the environment is fully known.
We apply \gls{potlp} to multiple tasks in simulation and the real world, showing improvement over a heuristic-driven baseline.

\section{Preliminaries}
\glsresetall

\noindent\textbf{\gls{lts}: }
We use an \gls{lts} to represent a discretized version of the robot's environment.
An LTS $\mathcal{T}$ is a tuple  $(X, x_0, \delta_\mathcal{T}, w, \Sigma, l)$, where: $X$ is a discrete set of states of the robot and the world, $x_0 \in X$ is the initial state, $\delta_\mathcal{T} \subseteq X \times X$ is the set of possible transitions between states, the weight $w: (x_i, x_j) \rightarrow \mathbb{R}^{+}$ is the cost incurred by making the transition $(x_i, x_j)$, $\Sigma$ is a set of \emph{propositions} with $\sigma \in \Sigma$, and the labeling function $l: X \rightarrow 2^\Sigma$ is used to assign which propositions are true in state $x \in X$.
An example \gls{lts} is shown in Fig.~\ref{fig:defining-actions}A where  $\Sigma=\{\texttt{fire}, \texttt{extinguisher}\}$, $l(x_3)=\{\texttt{extinguisher}\}$ and $l(x_4)=\emptyset$.
States $x~\in~X$ refer to physical locations, and labels like $\texttt{fire}$ or $\texttt{extinguisher}$ indicate whether there is a fire or extinguisher (or both) at that location.
A finite trajectory $\tau$ is a sequence of states $\tau = x_0 x_1 \ldots x_n$ where $(x_i, x_{i+1}) \in \delta_\mathcal{T}\ $ which generates a finite word $\omega= \omega_0 \omega_1 \ldots \omega_n$ where each letter $\omega_i = l(x_i)$.

\vspace{5pt}
\noindent\textbf{\gls{scltl}: }
To specify tasks, we use \gls{scltl}, a fragment of \gls{ltl}  \cite{Kupferman2001}.
Formulas are written over a set of atomic propositions $\Sigma$ with Boolean (negation ($\lnot$), conjunction ($\wedge$), disjunction ($\vee$)) and temporal (next ($\bigcirc$), until ($\mathcal{U}$), eventually ($\lozenge$)) operators.
The syntax is:
\begin{equation*}
    \varphi := \sigma \setor \lnot \sigma \setor \varphi \wedge \varphi \setor \varphi \vee \varphi  \setor \bigcirc \varphi \setor \varphi\ \mathcal{U}\ \varphi \setor \lozenge \varphi,
\end{equation*}
where $\sigma \in \Sigma$, and $\varphi$ is an \gls{scltl} formula.
The semantics are defined over infinite words $\omega^{\text{inf}} = \omega_0 \omega_1 \omega_2 \ldots$ with letters $\omega_i \in 2^\Sigma$.
Intuitively, $\bigcirc \varphi$ is satisfied by a given $\omega^{\text{inf}}$ at step $i$ if $\varphi$ is satisfied at the next step ($i+1$), $\varphi_1 \mathcal{U} \varphi_2$ is satisfied if $\varphi_1$ is satisfied at every step until $\varphi_2$ is satisfied, and $\lozenge \varphi$ is satisfied at step $i$ if $\varphi$ is satisfied at some step $j \geq i$.
For the complete semantics of \gls{scltl}, refer to \citet{Kupferman2001}. 
For example, $(\lnot \texttt{fire}\ \mathcal{U}\ \texttt{extinguisher}) \wedge \lozenge \texttt{fire}$ specifies the robot should avoid fire until reaching the extinguisher and eventually go to the fire.

\vspace{5pt}
\noindent \textbf{\gls{dfa}:}
A \gls{dfa}, constructed from an \gls{scltl} specification, is a tuple $\mathcal{D}_{\varphi} = (Z, z_0, \Sigma, \delta_\mathcal{D}, F)$.
A \gls{dfa} is composed of a set of states, $Z$, with an initial state $z_0 \in Z$.
The transition function $\delta_\mathcal{D}: Z \times 2^\Sigma \rightarrow Z$ takes in the current state, $z \in Z$, and a letter $\omega_i \in 2^\Sigma$, and returns the next state $z \in Z$ in the \gls{dfa}.
The \gls{dfa} has a set of accepting states, $F \subseteq Z$, such that if the execution of the \gls{dfa} on a finite word $\omega=\omega_0 \omega_1 \ldots \omega_n$ ends in a state $z \in F$, the word belongs to the language of the \gls{dfa}. 
While in general LTL formulas are evaluated over infinite words, the truth value of \gls{scltl} can be determined over finite traces. 
Fig.~\ref{fig:overview}A shows the \gls{dfa} for $(\lnot \texttt{fire}\ \mathcal{U}\ \texttt{extinguisher}) \wedge \lozenge \texttt{fire}$.

\vspace{5pt}
\noindent \textbf{\gls{pa}:}
A \gls{pa} is a tuple $(P, p_0, \delta_P, w_P, F_P)$ which captures this combined behavior of the \gls{dfa} and \gls{lts}.
The states $P = X \times Z$ keep track of both the \gls{lts} and \gls{dfa} states, where $p_0 = (x_0, z_0)$ is the initial state.
A transition between states is possible iff the transition is valid in both the \gls{lts} and \gls{dfa}, and is defined by $\delta_P = \{(p_i, p_j) \ |\ (x_i, x_j) \in \delta_\mathcal{T}, \delta_\mathcal{D}(z_i, l(x_j)) = z_j\}$.
When the robot moves to a new state $x \in X$ in the \gls{lts}, it transitions to a state in the \gls{dfa} based on the label $l(x)$. 
The weights on transitions in the \gls{pa} are the weights in the \gls{lts} $\left(w_P(p_i, p_j) = w(x_i, x_j)\right)$. 
Accepting states in the \gls{pa}, $F_P$, are those states with accepting \gls{dfa} states ($F_P = X \times F$).

\vspace{5pt}
\noindent \textbf{Partially/Locally Observable Markov Decision Process:}
We model the world as a \gls{lomdp}~\citep{merlin2020locally}---a \gls{pomdp}~\cite{pineau2002pomdps} where space within line of sight from our agent is fully observable---formulated as a belief MDP, and written as the tuple $(B, \mathcal{A}, P, R)$. 
In a belief MDP, $B$ represents the belief over states in an MDP, which in our case are the states from the \gls{pa} defined by a given task and environment. 
We can think of each belief state as distributions over three entities $b_t = \{ b_{\mathcal{T}}, b_x, b_z \}$: the environment (abstracted as an \gls{lts}) $b_\mathcal{T}$, the agent's position in that environment $b_x$, and the agent's progress through the \gls{dfa} defined by the task $b_z$.
With our \gls{lomdp} formulation, we collapse $b_x$ and $b_z$ to point estimates, and consider any states in $\mathcal{T}$ that have been seen by the robot as fully observable, allowing us to treat $b_\mathcal{T}$ as a labeled occupancy grid that is revealed as the robot explores. 
$\mathcal{A}$ is the set of available actions from a given state in the \gls{pa}, which at the lowest level of abstraction are the feasible transitions given by $\delta_P$\adam{PA transitions}. $P(b_{t+1} | b_t, a_t)$ and $R(b_{t+1}, b_t, a_t)$ are the transition probability and instantaneous cost, respectively, of taking an action $a_t$ to get from a belief $b_t$ to $b_{t+1}$.




\section{Planning in unknown environments with co-safe LTL specifications}
\label{sec:planning}

Our objective is to minimize the total expected cost of satisfying an \gls{scltl} specification in partially revealed environments.
Using the \gls{pomdp} model, we can represent the expected cost of taking an action using a belief-space variant of the Bellman equation \citep{pineau2002pomdps}:
\begin{equation}\label{eq:base-eq}
\begin{split}
Q(b_t, a_t) = \sum_{b_{t+1}} P(b_{t+1} | b_t, a_t) \Big[R(b_{t+1}, b_{t}, a_t) \\ 
+ \min_{a_{t+1}\in \mathcal{A}(b_{t+1})} Q(b_{t+1}, a_{t+1}) \Big],
\end{split}
\end{equation}
where $Q(b_t, a_t)$ is the belief-action cost of taking action $a_t$ given belief $b_t$. This equation can, in theory, be solved to find the optimal action for any belief. However, given the size of the environments we are interested in, directly updating the belief in evaluation of Eq.~\eqref{eq:base-eq} is intractable for two reasons.
First, due to the \emph{curse of dimensionality}, the size of the belief grows exponentially with the number of states. 
Second, by the \emph{curse of history}, the number of iterations needed to solve Eq.~\eqref{eq:base-eq} grows exponentially with the planning horizon.

\subsection{Defining the Set of High-Level Actions}
\label{sec:sub-action-set}

To overcome the challenges associated with solving Eq.~\eqref{eq:base-eq}, we first identify a set of discrete, \emph{high-level actions}, then build off an insight from \citet{Stein2018}: that we can simplify computing the outcome of each action by splitting future beliefs into two classes---futures where a given action is successful, and futures where it is not.
Note that this abstraction over the belief space does not exist in general, and is enabled by our assumption of perfect local perception.

\begin{figure}[t]
\vspace{2pt}
    \centering
    \includegraphics[width=\columnwidth]{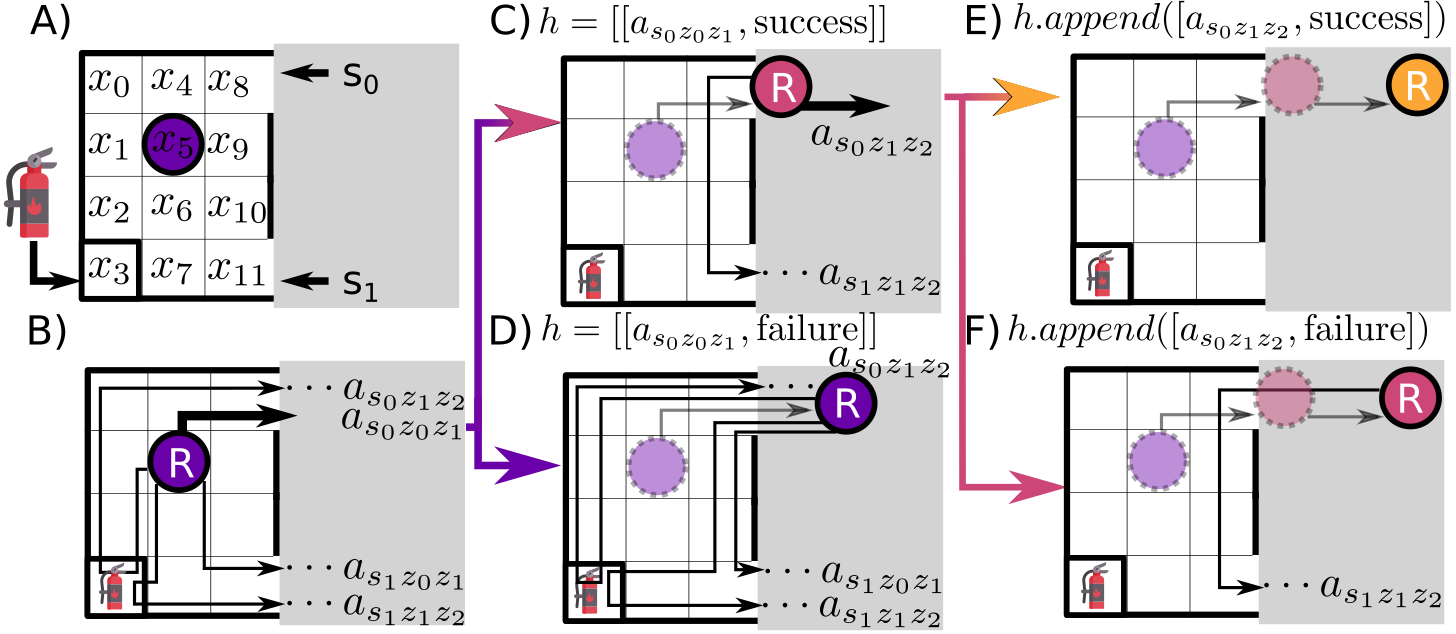}
    \vspace{-18pt}
    \caption{\textbf{High-level actions.} 
    The robot (``R'') is in a partially explored environment abstracted into an \gls{lts} with two subgoals $s_0$ and $s_1$ (A).
    As the robot considers high-level actions at different stages of its plan (B), its color indicates belief of the \gls{dfa} state (according to Fig.~\ref{fig:overview}A).
    In this rollout, the robot considers action $a_{s_0z_0z_1}$, and outcomes where it succeeds (C), and fails (D).
    If successful, the robot has two actions available and considers $a_{{s_0}{z_1}{z_2}}$, which in turn could succeed (E) or fail (F).
    }
    \label{fig:defining-actions}
    \vspace{-15pt}
\end{figure}

To satisfy a task specification, our agent must take actions to transition to an accepting state in the \gls{dfa}.
For example, given the specification $(\lnot \texttt{fire}\ \mathcal{U}\ \texttt{extinguisher})\wedge \lozenge \texttt{fire}$ as in Fig.\ \ref{fig:overview} and \ref{fig:defining-actions}, the robot must first retrieve the extinguisher and then reach the fire.
If the task cannot be completed entirely in the known space, the robot must act in areas that it has not yet explored.
As such, our action set is defined by \emph{subgoals} $\mathbb{S}$---each a point associated with a contiguous boundary between free and unexplored space---and the \gls{dfa}. 
Specifically, when executing an action, the robot first plans through the \gls{pa} in known space to a subgoal, and then enters the unexplored space beyond that subgoal to attempt to transition to a new state in the \gls{dfa}.

For belief state $b_t$, action $a_{sz'z''}$ defines the act of traveling from the current state in the \gls{lts} $x$ to reach subgoal $s \in \mathbb{S}_t$ at a \gls{dfa} state $z'$, and then attempting to transition to \gls{dfa} state $z''$ in unknown space. 
Our newly defined set of possible actions from belief state $b_t$ is: $\mathcal{A}(b_t) = \{a_{sz'z''}\ |\ s \in \mathbb{S}_t , z' \in Z_{\text{reach}}(b_t, s), z'' \in Z_{\text{next}}(z') \}$ where $Z_{\text{reach}}(b_t, s)$ is the set of \gls{dfa} states that can be reached while traveling in known space in the current belief $b_t$ to subgoal $s\in \mathbb{S}_t$, 
and $Z_{\text{next}}(z') = \{z'' \in Z\ |\ \exists \omega_i \in 2^\Sigma \textrm{ s.t. } z''=\delta_\mathcal{D}(z',\omega_i)\}$ is the set of \gls{dfa} states that can be visited in one transition from $z'$.
Fig.~\ref{fig:defining-actions}B illustrates an example of available high-level actions.

When executing action $a_{sz'z''}$, the robot reaches subgoal $s$ in \gls{dfa} state $z'$, accumulating a cost $D(b_t, a_{sz'z''})$ which is computed using Dijkstra's algorithm in the known map.
Once the robot enters the unknown space beyond the subgoal, the action has some probability $P_S$ of successfully transitioning from $z'$ to $z''$, denoted as: ${P_{S}(b_t, a_{sz'z''}) \equiv \sum_{b_{t+1}} P(b_{t+1} | a_{sz'z''}, b_t)\mathbb{I}[Z(b_{t+1})=z'']}$, where $Z(b_{t+1})$ refers to $b_z$ at the next time step and $\mathbb{I}[Z(b_{t+1})=z'']$ is an indicator function for belief states where the agent has reached the \gls{dfa} state $z''$.
Each action has an expected cost of success $R_{S}(b_t, a_{sz'z''})$ such that $R_{S}(b_t, a_{sz'z''}) + D(b_t, a_{sz'z''}) \equiv \frac{1}{P_{S}} \sum_{b_{t+1}} P(b_{t+1} | b_t, a_{sz'z''}) R(b_{t+1}, b_t, a_{sz'z''}) \mathbb{I}[Z(b_{t+1}) = z'']$ and expected cost of failure $R_{F}(b_t, a_{sz'z''})$, which is equivalently defined for $Z(b_{t+1}) \neq z''$ and normalized with $\frac{1}{1-P_{S}}$. 
By estimating these values via learning (discussed in Sec.~\ref{sec:learning}), we can express the expected instantaneous cost of an action as $\sum_{b_{t+1}} P(b_{t+1} | b_t, a_{sz'z''}) R(b_{t+1}, b_{t}, a_{sz'z''}) = D + P_{S} R_{S} + (1-P_{S})R_{F}$.


\begin{figure*}[ht]
\centering
\begin{minipage}{.475\textwidth}
    \vspace{-5pt}
    \begin{algorithm}[H]
    \SetAlgoLined
    \SetFuncSty{textsc}
    \SetCommentSty{text}
    \DontPrintSemicolon
    
    \SetKwProg{Fn}{Function}{:}{}
    \SetKwFunction{FMain}{PO-TLP}
    \SetKwFunction{FHighLevelPlan}{HighLevelPlan}
    \SetKwFunction{FUpdateSubgoals}{UpdateSubgoals}
    \SetKwFunction{FGetPossibleActions}{GetPossibleActions}
    \SetKwFunction{FActAndObserve}{ActAndObserve}
    \SetKwFunction{FEstActionProps}{EstActProps}
    \SetKwFunction{FSample}{Sample}
    \SetKwFunction{FRollout}{Rollout}
    
    \Fn(\tcp*[f]{$\theta$: Network parameters}){\FMain{$\theta$}}{
        $b \leftarrow \{b_{\mathcal{T}_0}, b_{x_0}, b_{z_0}\},\ \textit{Img} \leftarrow \textit{Img}_0$\;
    
        \While{True}{
            \If(\tcp*[f]{In accepting state}){$b_z \in F$} {
                \KwRet {SUCCESS}
            }
            \vspace{-5pt}
            \begin{flushleft}
            $a_{sz'z''}^* \leftarrow \textsc{HighLevelPlan}(b, \textit{Img}, \theta, \mathcal{A}(b))$ \;
            $b,\,\textit{Img} \leftarrow $ \FActAndObserve{{$a_{sz'z''}^*$}} \;
            \vspace{-5pt}
            \end{flushleft}
        }
    }
    \begin{flushleft}
    \Fn{\FHighLevelPlan{$b$, $\text{Img}$, $\theta$, $\mathcal{A}(b)$}}{
        \For{$a \in \mathcal{A}(b)$}{
            $\mathunderline[softblue]{a.P_S}, \mathunderline[softblue]{a.R_S}, \mathunderline[softblue]{a.R_F} \leftarrow \textsc{EstProps}(b, a, \textit{Img}\, \rvert \,\theta)$\;
        }
    
        
        $a_{sz'z''}^*  \leftarrow \textsc{PO-UCT}(b, \mathcal{A}(b))$ \; 
        \begin{flushleft}
            \vspace{-5pt}
            \KwRet {$a_{sz'z''}^*$}
            \vspace{-5pt}
        \end{flushleft}
    }
    \vspace{-6pt}
    \end{flushleft}   
    \caption{PO-TLP}
    \label{alg:po-tlp}
    \end{algorithm}
\end{minipage}%
~~~~~~~
\begin{minipage}{.49\textwidth}
  \centering
    \vspace{5pt}
    \centering
    \includegraphics[width=1.0\columnwidth]{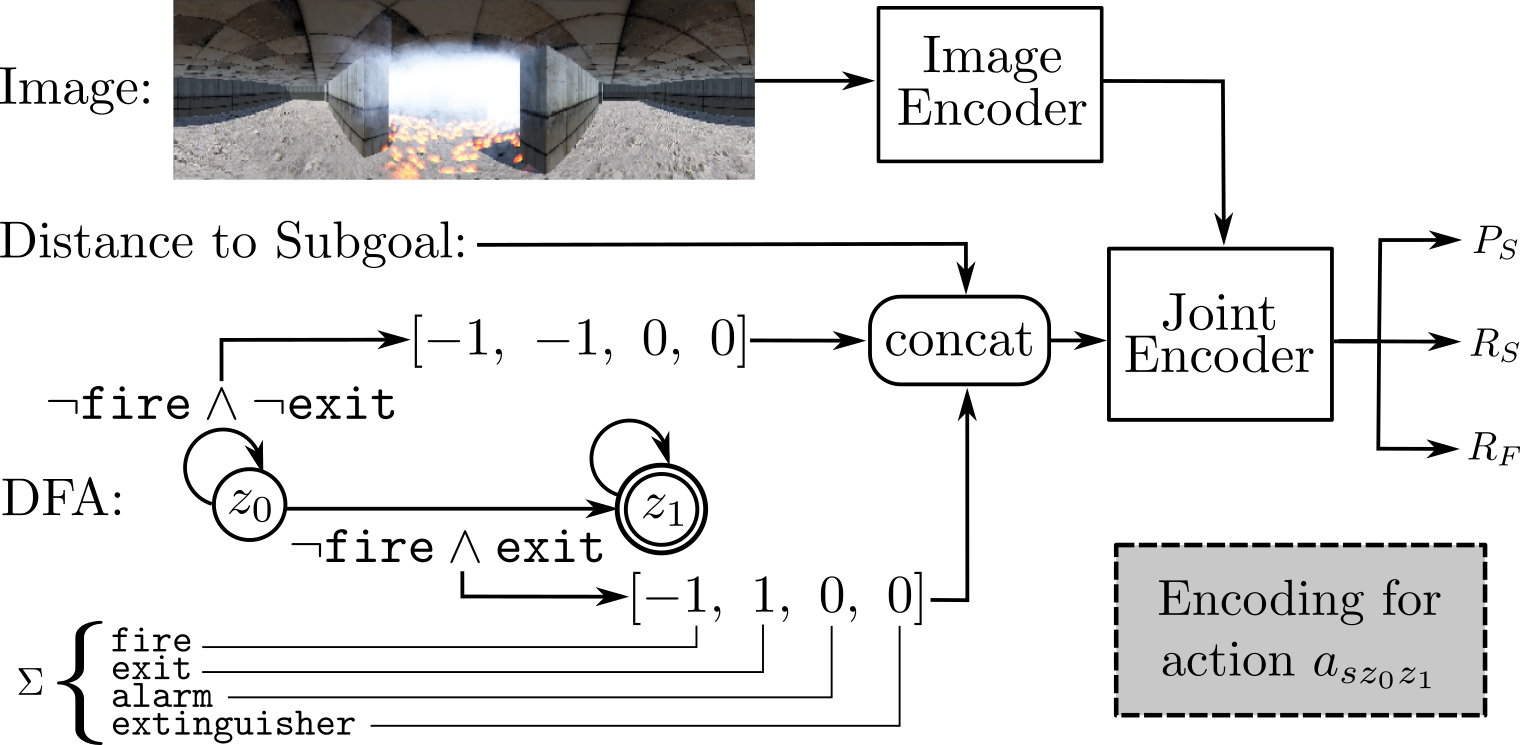}
    \caption{\textbf{Neural network inputs and outputs.} Estimating $P_S$, $R_F$, and $R_S$ for action $a_{sz_0z_1}$: attempting to reach an exit while avoiding fire.
    }
    \label{fig:network-arch}
\end{minipage}
\vspace{-15pt}
\end{figure*}

\subsection{Estimating Future Cost using High-Level Actions}
\label{sec:sub-planning-with-actions}

\begin{figure*}[b]
\vspace{-15pt}
\begin{equation}
\tag{3}
\label{eq:ltl-subgoal-planning-equation}
  \begin{split}
  Q(b_h, a_{sz'z''}) =  D(b_h, a_{sz'z''}) +
  \mathunderline[softblue]{P_S(b_h, a_{sz'z''})} \times
  \left[\mathunderline[softblue]{R_S(b_h, a_{sz'z''})} + \min_{a_S \in \mathcal{A}(b_{h_S})} Q(b_{h_S}, a_S)\right] \\
  \mathunderline[softblue]{\small\parbox{10em}{\emph{Underline denotes terms}\\[-2pt]\emph{we estimate via learning}\\[-10pt]}} \qquad \qquad \qquad + \left[1-\mathunderline[softblue]{P_S(b_h, a_{sz'z''})}\right] \times
  \left[\mathunderline[softblue]{R_F(b_h, a_{sz'z''})} + \min_{a_F \in \mathcal{A}(b_{h_F})} Q(b_{h_F}, a_F) \right]
  \end{split}\raisetag{1\baselineskip}
\end{equation}
\vspace{-5pt}
\end{figure*}

We now examine planning to satisfy a specification with minimum cost using these actions, remembering to consider both the case where an action succeeds---the robot transitions to the desired state in the \gls{dfa}---and when it fails. 
We refer to a complete simulated trial as a \emph{rollout}. 
Since we cannot tractably update and maintain a distribution over maps during a rollout, we instead keep track of the \emph{rollout history} $h = [[a_0, o_0], \ldots, [a_n, o_n]]$ (a sequence of high-level actions $a_i$ considered during planning and their simulated respective outcomes $o_i = \{\texttt{success}, \texttt{failure}\}$).
Recall that, during planning, we assume that the agent knows its position $x$ in known space $\mathcal{T}_\text{known}$.
Additionally, conditioned on whether an action $a_{sz'z''}$ is simulated to succeed or fail, we assume no uncertainty over the resulting \gls{dfa} state, collapsing $b_z$ to $z''$ or $z'$, respectively.
Therefore, for a given rollout history, the agent knows its position in known space and its state in the \gls{dfa}. 
\added{These assumptions, while not suited to solving \gls{pomdp}s in general, fit within the \gls{lomdp} model.}


During a rollout, the set of available future actions is informed by actions and outcomes already considered in $h$.
For example, if we simulate an action in a rollout, and it fails, we should not consider that action a second time.
Conversely, we know a successful action would succeed if simulated again in that rollout.
In Fig.~\ref{fig:defining-actions}F, the robot imagines its first action $a_{s_0z_0z_1}$ succeeds, while its next action $a_{s_0z_1z_2}$ fails, making the rollout history $h=[[a_{s_0z_0z_1}, \texttt{success}], [a_{s_0z_1z_2}, \texttt{failure}]]$.
When considering the next step of this rollout, the robot knows it can always find an extinguisher beyond $s_0$, and there is no fire beyond $s_0$.
To track this information during planning, we define a \emph{rollout history-augmented belief} $b_h~=~\{\mathcal{T}_\text{known}, x, z, h\}$, which augments the belief with the actions and outcomes of the rollout up to that point. 
To reiterate, we maintain the history-augmented belief $b_h$ only during planning, to avoid the complexity of maintaining a distribution over possible future maps from possible future observations during rollout.

Using $b_h$, we also define a \emph{rollout history-augmented action set} $\mathcal{A}(b_h)$, in which actions known to be impossible based on $h$ are pruned, and with it, a \emph{rollout history-augmented success probability} $P_S(b_h, a_{sz'z''})$ which is identically one for actions known to succeed.
Furthermore, because high-level actions involve entering unknown space, instead of explicitly considering the distribution over possible robot states, we define a \emph{rollout history-augmented distance} function $D(b_h, a_{sz'z''})$, which takes into account physical location as a result of taking the last action in $b_h$.
If an action leads to a new subgoal ($s_{t+1} \neq s_t$), the agent accumulates the success cost $R_S$ of the previous action if that action was simulated to succeed and the failure cost $R_F$ if it was not.

By planning with $b_h$, the future expected reward can be written so that it no longer directly depends on the full future belief state $b_{t+1}$, allowing us to approximate it as follows:
\begin{equation}
\begin{split}
\sum_{b_{t+1}}P(&b_{t+1} | b_{t}, a_{sz'z''}) \min_{a' \in \mathcal{A}(b_{t+1})} Q(b_{t+1}, a') \approx \\
& P_S(b_h, a_{sz'z''}) \min_{a_S \in \mathcal{A}(b_{h_S})} Q(b_{h_S}, a_S) + \\
& [1-P_S(b_h, a_{sz'z''})] \min_{a_F \in \mathcal{A}(b_{h_F})} Q(b_{h_F}, a_F),
\end{split}
\end{equation}
where $h_S = h.\texttt{append}([a_{sz'z''}, \texttt{success}])$ is the rollout history conditioned on a successful outcome (and $h_F$ is defined similarly for failed outcomes).



\begin{figure*}[t]
\vspace{10pt}
\centering
\includegraphics[width=\textwidth]{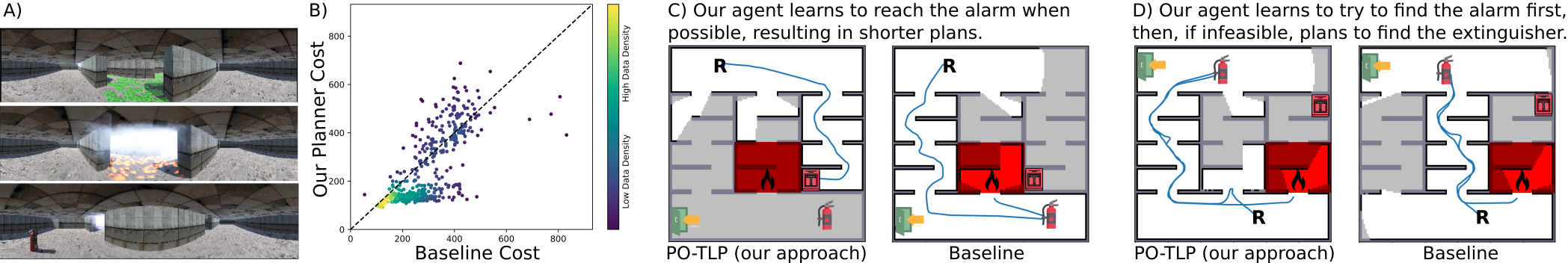}
\vspace{-18pt}
\caption{
A comparison between our planner and the baseline for 500 simulated trials in the Firefighting environment with the specification $(\lnot \texttt{fire}\ \mathcal{U}\ \texttt{alarm}) \vee ((\lnot \texttt{fire}\ \mathcal{U}\ \texttt{extinguisher}) \wedge \lozenge \texttt{fire})$.
The robot (``R'') learns to associate green tiling with the alarm, and hallways emanating white smoke with fire (A), leading to a 15\% improvement for total net cost for this task (B).
Our agent learns it is often advantageous to search for the alarm (C), so in cases where the alarm is reachable, we generally outperform the baseline (highlighted by the cluster of points in the lower half of B).
Our method is occasionally outperformed when the alarm can't be reached (D), though we perform better in the aggregate and always satisfy the specification.
}
\label{fig:fire_fighter_results}
\vspace{-15pt}
\end{figure*}


Our high-level actions and rollout history allow us to approximate Eq.~\eqref{eq:base-eq} with Eq.~\eqref{eq:ltl-subgoal-planning-equation}, \added{a finite horizon problem.}
Given estimates of $P_S$, $R_S$, and $R_F$---which are learned as discussed in Sec.~\ref{sec:learning}---we avoid explicitly summing over the distribution of all possible \gls{pa}s, thereby reducing the computational complexity of solving for the best action.



\subsection{Planning with High-Level Actions using PO-UCT}
\label{sub:mcts}
Eq.~\eqref{eq:ltl-subgoal-planning-equation} demonstrates how expected cost can be computed exactly using our high-level actions, \added{given estimated values of $P_S$, $R_S$, and $R_F$}.
However, considering every possible ordering of actions as in Eq.~\eqref{eq:ltl-subgoal-planning-equation} still involves significant computational effort\added{---exponential in both the number of subgoals and the size of the \gls{dfa}}.
Instead, we adapt \gls{pouct}\cite{POMCP}, a generalization of \gls{mcts} which tracks histories of actions and outcomes, to select the best action for a given belief using sampling.
The nodes of our search tree correspond to belief states $b_\hist$, and actions available at each node are defined according to the rollout history as discussed in Sec.~\ref{sec:sub-planning-with-actions}.
At each iteration, from this set we sample an action and its outcome according to the Bernoulli distribution parameterized by $P_S$, and accrue cost by $R_S$ or $R_F$.

This approach prioritizes the most promising branches of the search space, avoiding the cost of enumerating all states. 
By virtue of being an anytime algorithm, \gls{pouct} also enables budgeting computation time, allowing for faster online execution as needed on a real robot.
Once our agent has searched for the action with lowest expected cost, \added{$a_{sz'z''}^*$,} it generates a motion plan through space to the subgoal associated with the chosen action.
While moving along this path, the agent receives new observations of the world, updates its map, and replans (see Fig.~\ref{fig:overview} and Algorithm~\ref{alg:po-tlp}).


\section{Learning Transition Probabilities and Costs}
\label{sec:learning}

To plan with our high-level actions, we rely on values for $P_S$, $R_S$, and $R_F$ for arbitrary actions $a_{sz'z''}$. 
Computing these values explicitly from the belief (as defined in  Sec.~\ref{sec:sub-action-set}) is intractable, so we train a neural network to estimate them from visual input and an encoding of the action.

\subsection{Encoding a Transition}
\label{sub:encoding}

A successful action $a_{sz'z''}$ results in the desired transition in the \gls{dfa} from $z'$ to $z''$ occurring in unknown space.
However, encoding actions directly using \gls{dfa} states prevents our network from generalizing to other specifications without retraining.
Instead, we use an encoding that represents formulas over the set of propositions $\Sigma$ in negative normal form \cite{Li2013} over the truth values of $\Sigma$, which generalizes to any specification written with $\Sigma$ in similar environments.

To progress from $z'$ to $z''$ in unknown space, the robot must travel such that it remains in $z'$ until it realizes changes in proposition values that allow it to transition to $z''$.
We therefore define two $n$-element feature vectors $[\phi(z',z'), \phi(z',z'')]$ where $\phi \in \{-1, 0, 1\}^n$, which serve as input to our neural network.
For the agent to stay in $z'$, if the $i$th element in $\phi(z',z')$ is $1$, the corresponding proposition must be true at all times; if it is $-1$, the proposition must be false; and if it is $0$, the proposition has no effect on the desired transition. 
The values in $\phi(z',z'')$ are defined similarly for the agent to transition from $z'$ to $z''$. 
Fig. \ref{fig:network-arch} illustrates this feature vector for a task specification example.

\subsection{Network Architecture and Training}
\label{sub:architecture}
Our network takes as input a $128\!\times\!512$ RGB panoramic image centered on a subgoal, the scalar distance to that subgoal, and the two $n$-element feature vectors $\phi$ describing the transition of interest, as defined in Sec.~\ref{sub:encoding}.
The input image is first passed through 4 convolutional layers, after which we concatenate the feature vectors and the distance to the subgoal to each element (augmenting the number of channels accordingly), and continue encoding for 8 more convolutional layers.
Finally, the encoded features are passed through 5 fully connected layers, and the network outputs the properties required for Eq.~\eqref{eq:ltl-subgoal-planning-equation}---$P_S$, $R_S$, and $R_F$.
We train our network with the same loss function as in \citep{Stein2018}.


To collect training data, we navigate through environments using an autonomous, heuristic-driven agent in simulation, and teleoperation in the real world. 
We assume the agent has knowledge of propositions in its environments, so it can generate the feature vectors that encode actions for subgoals it encounters.
As the robot travels, we periodically collect images and the true values of $P_S$ (either 0 or 1), $R_S$, and $R_F$ for each potential action from the underlying map.


\begin{figure}[b]
\vspace{-10pt}
\centering
\includegraphics[width=1.0\columnwidth]{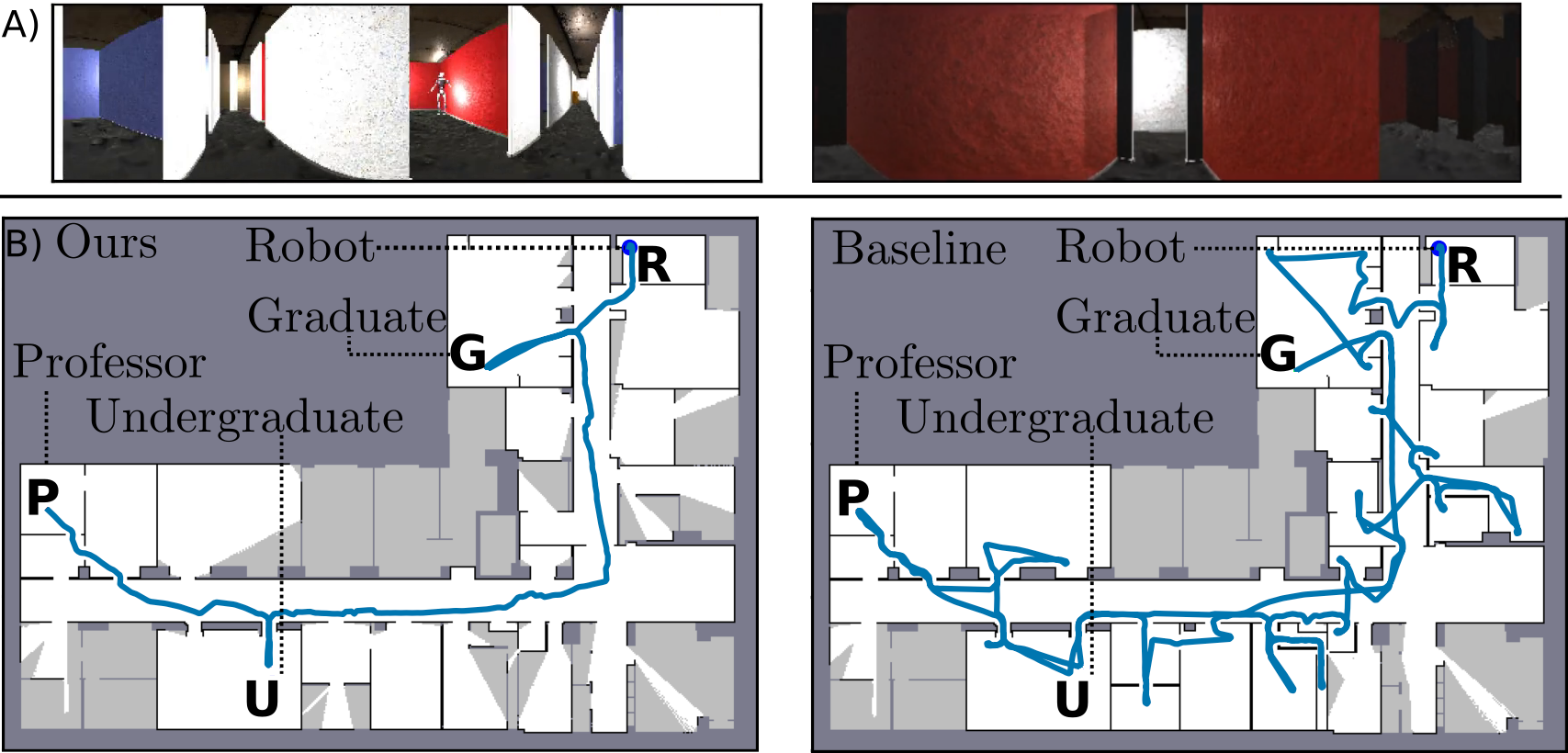}
\vspace{-18pt}
\caption{A) Visual scenes from our Delivery scenario in simulation. 
The rooms which can contain professors, graduate students, and undergraduates are colored differently and illuminated when occupied.
B) A comparison between our approach (left) and the baseline (right) for one of several simulated trials for the task $\lozenge \texttt{professor} \wedge \lozenge \texttt{grad} \wedge \lozenge \texttt{undergrad}$.
}
\label{fig:delivery_sim_results}
\vspace{-3pt}
\end{figure}

\section{Experiments}
\label{sec:results}




We perform experiments in simulated and real-world environments, comparing our approach with a baseline derived from \citet{Ayala2013temporal}.
The baseline solves similar planning-under-uncertainty problems using boundaries between free and unknown space to define actions, albeit with a non-learned selection procedure and no visual input.
Specifically, we compare the total distances traveled using each method.



\begin{figure*}[t]
\vspace{5pt}
\centering
\includegraphics[width=0.99\textwidth]{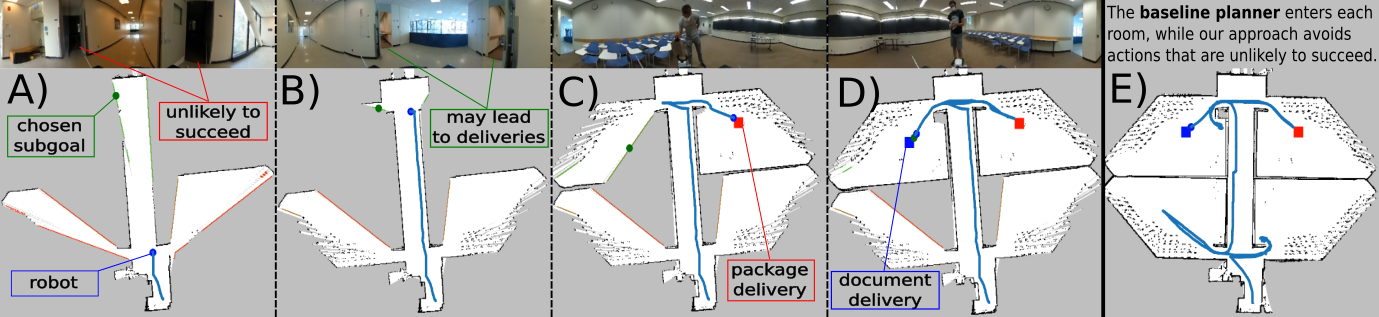}
\vspace{-8pt}
\caption{A comparison between our approach (A-D) and the baseline (E) for our real-world Delivery scenario.
Our agent (blue dot) correctly predicts actions likely to fail (e.g., dark rooms in A) and succeed (e.g., completing a delivery in an illuminated room in B).
Once the robot identifies delivery actions that lead to \gls{dfa} transitions in known space, it executes them (C-D).
Conversely, the baseline fully explores space before completing the task (E).
}
\label{fig:delivery_real_world_results}
\vspace{-15pt}
\end{figure*}

\subsection{Firefighting Scenario Results}
\label{sub:fire_fighter_results}

Our first environment is based on the firefighting robot example, simulated with the Unity game engine \citep{unity} and shown in Fig.~\ref{fig:fire_fighter_results}.
The robot is randomly positioned in one of two rooms, and the extinguisher and exit in the other.
One of three hallways connecting the rooms is randomly chosen to be a dead end with an alarm at the end of it, and is visually highlighted by a green tiled floor.
A hallway (possibly the same one) is chosen at random to contain a fire, which blocks passage and emanates white smoke.
Our network learns to associate the visual signals of green tiles and smoke to the hallways containing the alarm and the fire, respectively.

We run four different task specifications---using the same network without retraining to demonstrate its reusability:
\begin{flushenumerate}
\item $(\lnot \texttt{fire}\ \mathcal{U}\ \texttt{alarm}) \vee ((\lnot \texttt{fire}\ \mathcal{U}\ \texttt{extinguisher}) \wedge \lozenge \texttt{fire})$: avoid the fire until the alarm is found, or avoid the fire until the extinguisher is found, then find the fire.
\item $\lnot \texttt{fire}\ \mathcal{U}\ (\texttt{alarm} \wedge \lozenge \texttt{exit})$: avoid the fire until the alarm is found, then exit the building.
\item $(\lnot \texttt{fire}\ \mathcal{U}\ \texttt{extinguisher}) \wedge \lozenge (\texttt{fire} \wedge \lozenge \texttt{exit})$: avoid the fire until the extinguisher is found, then put out the fire and exit the building.
\item $\lnot \texttt{fire}\ \mathcal{U}\ \texttt{exit}$: avoid the fire and exit the building.
\end{flushenumerate}

Over {\raise.17ex\hbox{$\scriptstyle\sim$}}3000 trials across different simulated environments and these four specifications, we demonstrate improvement for our planner over the baseline (see Table~\ref{tab:results}).
Fig.~\ref{fig:fire_fighter_results} gives a more in depth analysis of the results for specification 1. 

\begin{table}[h]
  \vspace{-2pt}
  \begin{center}
    \caption{\vspace{-3pt}}
    \label{tab:results}
    \vspace*{0mm}
    \begin{tabular}{c|c|c|c|c|c} 
    \multicolumn{1}{c|}{\textbf{}} & 
    \multicolumn{3}{|c|}{\textbf{Average Cost}} & 
    \multicolumn{2}{|c}{\textbf{Percent Savings}} \\
    \cline{2-6}    
      \textbf{Spec} & 
      \textbf{\!\!\!Known Map\!\!\!} &  \textbf{\!\!\!Baseline\!\!\!} & \textbf{\!\!\!Ours\!\!\!} & \textbf{\!\!\!Net Cost\!\!\!} & \textbf{\!\!\!Per Trial (S.E.)\!\!\!}  \\ %
      \hline
      1 &  187.0 & 264.6 & 226.9 & 15\% & 14.4\% (1.3)\\ 
      2 & 392.1 & 696.0 & 461.2 & 34\% & 28.4\% (1.0)\\ 
      3 & 364.3 & 539.3 & 471.3 & 13\% & 6.1\% (1.3)\\ 
      4 & 171.2 & 269.1 & 203.3 & 25\% & 14.6\% (1.0)\\ 
    \end{tabular}
  \end{center}
  \vspace{-15pt}
\end{table}

\subsection{Delivery Scenario Results in Simulation}
\label{sub:mit-floorplan}


We scale our approach to larger simulated environments using a corpus of academic buildings containing \emph{labs}, \emph{classrooms}, and \emph{offices}, all connected by hallways.
Our agent must deliver packages to three randomly placed individuals in these environments: one each of a professor, graduate student, and undergraduate, using the specification $\lozenge \texttt{professor} \wedge \lozenge \texttt{grad} \wedge \lozenge \texttt{undergrad}$.
Professors are located randomly in offices, graduate students in labs, and undergraduates in classrooms, which have differently colored walls in simulation. 
Rooms that are occupied have the lights on whereas other rooms are not illuminated.

Our agent is able to learn from these visual cues to navigate more efficiently.
Over 80 simulations across 10 environments, the mean per-trial cost improvement of our agent compared with the baseline is 13.5\% (with 6.1\% standard error). Our net cost savings, summed over all trials, is 7.8\%.
Fig.~\ref{fig:delivery_sim_results} illustrates our results.

\subsection{Delivery Scenario Results in the Real World}
\label{sub:real-world-results}

We extend our Delivery scenario to the real world using a Toyota Human Support Robot (HSR) \cite{Yamamoto2019} with a head-mounted panoramic camera \cite{ThetaCamera} in environments with multiple rooms connected by a hallway.
The robot must deliver a package and a document to two people, either unordered $(\lozenge \texttt{DeliverDocument} \wedge \lozenge \texttt{DeliverPackage})$ or in order $(\lozenge (\texttt{DeliverDocument} \wedge \lozenge \texttt{DeliverPackage}))$.
As in simulation, rooms are illuminated only if occupied.

We ran 5 trials for each planner spanning both specifications and 3 different target positions in a test environment different from the one used to collect training data.
We show improved performance over the baseline in all cases with a mean per-trial cost improvement of 36.6\% (6.2\% standard error), and net cost savings, summed over all trials, of 36.5\%.
As shown in Fig.~\ref{fig:delivery_real_world_results}, the baseline enters the nearest room regardless of external signal, while our approach prioritizes illuminated rooms, which are more likely to contain people.


\section{Related Works}


Temporal logic synthesis has been used to generate provably correct controllers, although predominantly in fully known environments \citep{Kress-Gazit2018, Fainekos2009, Bhatia2010, Lacerda2014, Smith2011}.
\added{Recent work has looked at satisfying \gls{ltl} specifications under uncertainty beyond LOMDPs \citep{bouton2020point, ahmadi2020stochastic}, yet these works are restricted to small state spaces due to the general nature of the \gls{pomdp}s they handle.}
\added{Other work has explored more restricted sources of uncertainty, such as scenarios where tasks specified in \gls{ltl} need to be completed in partially explored environments \citep{Ayala2013temporal, Sarid2012guaranteeing, Lahijanian2016iterative, guo2015multi, guo2013revising} or by robots with uncertainty in sensing, actuation, or the location of other agents \citep{vasile2016control, kantaros2020reacttlp, kantaros2019optimal, svorenova2015temporal}.}
\added{When these robots plan in partially explored environments, they take the best possible action given the known map, but either ignore or make naive assumptions about unknown space; conversely, we use learning to incorporate information about unexplored space.}
\nocite{horak2019solving}


To minimize the cost of satisfying \gls{ltl} specifications, other recent works have used learning-based approaches \cite{littman2017ltl, Fu2015probably, li2019formal, icarte2018teaching}, yet these methods are limited to relatively small, fully observable environments. 
\citet{sadigh2014learning} and \citet{Fu2015probably} apply learning to unknown environments and learn the transition matrices for \acrshort{mdp}s built from \gls{ltl} specifications.
However, these learned models do not generalize to new specifications and are demonstrated on relatively small grid worlds.
\citet{paxton2017combining} introduce uncertainty during planning, but limit their planning horizon to 10 seconds, which is insufficient for the specifications explored here.
\added{\citet{carr2020verifiable} synthesize a controller for verifiable planning in \gls{pomdp}s using a recurrent neural network, yet are limited to planning in small grid worlds.}
\section{Conclusion and Future Work}

In this work, we present a novel approach to planning to solve \gls{scltl} tasks in partially revealed environments. 
Our approach learns from raw sensor information and generalizes to new environments and task specifications without the need to retrain.
We hope to extend our methods to real-world planning domains that involve manipulation and navigation.






\printbibliography

@inproceedings{Stein2018,
author = {Stein, Gregory J and Bradley, Christopher and Roy, Nicholas},
booktitle = {CoRL},
title = {{Learning over Subgoals for Efficient Navigation of Structured, Unknown Environments}},
year = {2018}
}

@inproceedings{Ayala2013temporal,
author = {Ayala, A. I. Medina and Andersson, S. B. and Belta, C.},
booktitle = {IROS},
title = {{Temporal logic motion planning in unknown environments}},
year = {2013}
}

@inproceedings{Sarid2012guaranteeing,
author = {Sarid, Shahar and Xu, Bingxin and Kress-Gazit, Hadas},
booktitle = {RSS},
title = {{Guaranteeing High-Level Behaviors While Exploring Partially Known Maps}},
year = {2012}
}

@article{Lahijanian2016iterative,
author = {Lahijanian, Morteza and Maly, Matthew R. and Fried, Dror and Kavraki, Lydia E. and Kress-Gazit, Hadas and Vardi, Moshe Y.},
journal = {TRO},
title = {{Iterative Temporal Planning in Uncertain Environments With Partial Satisfaction Guarantees}},
year = {2016}
}

@INPROCEEDINGS{Lacerda2014,
author={B. {Lacerda} and D. {Parker} and N. {Hawes}},
booktitle={IROS},
title={Optimal and dynamic planning for Markov decision processes with co-safe LTL specifications},
year={2014},
}

@inproceedings{Fu2015probably,
author = {Fu, Jie and Topcu, Ufuk},
booktitle = {RSS},
title = {{Probably Approximately Correct MDP Learning and Control With Temporal Logic Constraints}},
year = {2015}
}

@inproceedings{sadigh2014learning,
  title={A learning based approach to control synthesis of markov decision processes for linear temporal logic specifications},
  author={Sadigh, Dorsa and Kim, Eric S and Coogan, Samuel and Sastry, S Shankar and Seshia, Sanjit A},
  booktitle={CDC},
  year={2014},
}

@Inbook{Li2013,
author="Li, Jianwen
and Pu, Geguang
and Zhang, Lijun
and Wang, Zheng
and He, Jifeng
and Guldstrand Larsen, Kim",
title="On the Relationship between LTL Normal Forms and B{\"u}chi Automata",
bookTitle="Theories of Programming and Formal Methods",
year={2013},
}

@article{Kaelbling:1998:PAP:1643275.1643301,
author = {Kaelbling, Leslie Pack and Littman, Michael L. and Cassandra, Anthony R.},
title = {Planning and Acting in Partially Observable Stochastic Domains},
journal = {Artificial Intelligence},
year = {1998},
}

@incollection{Littman1995362,
title = "Learning policies for partially observable environments: Scaling up ",
booktitle = "Machine Learning Proceedings",
year = "1995",
author = "Michael L. Littman and Anthony R. Cassandra and Leslie Pack Kaelbling",
}

@incollection{POMCP,
title = {Monte-Carlo Planning in Large POMDPs},
author = {Silver, David and Veness, Joel},
booktitle = {Advances in Neural Information Processing Systems},
year = {2010},
}

@inproceedings{icarte2018teaching,
author = {Toro Icarte, Rodrigo and Klassen, Toryn Q. and Valenzano, Richard and McIlraith, Sheila A.},
title = {Teaching Multiple Tasks to an RL Agent Using LTL},
year = {2018},
booktitle = {AAMAS},
}

@article{littman2017ltl,
  title={Environment-independent task specifications via GLTL},
  author={Littman, Michael L and Topcu, Ufuk and Fu, Jie and Isbell, Charles and Wen, Min and MacGlashan, James},
  journal={arXiv},
  year={2017}
}

@inproceedings{paxton2017combining,
  title={Combining neural networks and tree search for task and motion planning in challenging environments},
  author={Paxton, Chris and Raman, Vasumathi and Hager, Gregory D and Kobilarov, Marin},
  booktitle={IROS},
  year={2017},
}

@article{li2019formal,
  title={A formal methods approach to interpretable reinforcement learning for robotic planning},
  author={Li, Xiao and Serlin, Zachary and Yang, Guang and Belta, Calin},
  journal={Science Robotics},
  year={2019},
  publisher={Science Robotics}
}

@article{Kress-Gazit2018,
journal = {Annual Review of Control, Robotics, and Autonomous Systems},
author={Kress-Gazit, Hadas and Lahijanian, Morteza and Raman, Vasumathi},
title = {{Synthesis for Robots: Guarantees and Feedback for Robot Behavior}},
year = {2018}
}

@inproceedings{Kupferman2001,
author = {Kupferman, Orna and Vardi, Moshe Y.},
title = {{Model checking of safety properties}},
booktitle = {Formal Methods in System Design},
year = {2001},
}

@inproceedings{Bhatia2010,
author = {Bhatia, Amit and Kavraki, Lydia E. and Vardi, Moshe Y.},
booktitle = {ICRA},
title = {{Sampling-based motion planning with temporal goals}},
year = {2010}
}

@article{Smith2011,
author = {Smith, Stephen L and T{\^{u}}mov{\'{a}}, Jana and Belta, Calin and Rus, Daniela},
journal = {IJRR},
number = {14},
publisher = {SAGE PublicationsSage UK: London, England},
title = {{Optimal path planning for surveillance with temporal-logic constraints}},
volume = {30},
year = {2011}
}

@article{Fainekos2009,
author = {Fainekos, Georgios E. and Girard, Antoine and Kress-Gazit, Hadas and Pappas, George J.},
journal = {Automatica},
title = {{Temporal logic motion planning for dynamic robots}},
year = {2009}
}

@article{Yamamoto2019,
author = {Yamamoto, Takashi and Terada, Koji and Ochiai, Akiyoshi and Saito, Fuminori and Asahara, Yoshiaki and Murase, Kazuto},
title = {{Development of HSR as the research platform of a domestic mobile manipulator}},
journal = {ROBOMECH}, 
year = {2019}
}

@misc{unity,
  author = {{Unity Technologies}},
  title = {Unity Game Engine},
  howpublished = {\url{unity3d.com}},
  year = 2019
}

@inproceedings{spot,
  author = {Alexandre Duret-Lutz and Alexandre Lewkowicz and Amaury
		  Fauchille and Thibaud Michaud and Etienne Renault and
		  Laurent Xu},
  title = {Spot 2.0 --- a framework for {LTL} and $\omega$-automata
		  manipulation},
  booktitle = {ATVA},
  publisher = {Springer},
  year = {2016},
}

@techreport{pineau2002pomdps,
  author =	 {Joelle Pineau and Sebastian Thrun},
  title =	 {An integrated approach to hierarchy and abstraction for
                  {POMDPs}},
  institution =	 {CMU},
  year =	 2002,
}

@misc{ThetaCamera,
  author = { },
  title = {{Ricoh Theta S Panoramic Camera}},
  year = {2019},
}

@inproceedings{kantaros2020reacttlp,
title={Reactive Temporal Logic Planning for Multiple Robots in Unknown Environments},
  author={Kantaros, Yiannis and Malencia, Matthew and Kumar, Vijay and Pappas, George J},
booktitle = {ICRA},
year = {2020}
}

@inproceedings{kantaros2019optimal,
  title={Optimal Temporal Logic Planning for Multi-Robot Systems in Uncertain Semantic Maps},
  author={Kantaros, Yiannis and Pappas, George},
  booktitle={IROS},
  year={2019}
}

@inproceedings{vasile2016control,
  title={Control in Belief Space with Temporal Logic Specifications},
  author={Vasile, Cristian-Ioan and Leahy, Kevin and Cristofalo, Eric and Jones, Austin and Schwager, Mac and Belta, Calin},
  booktitle={CDC},
  year={2016}
}

@inproceedings{carr2020verifiable,
  title={Verifiable RNN-Based Policies for POMDPs Under Temporal Logic Constraints},
  author={Carr, Steven and Jansen, Nils and Topcu, Ufuk},
  booktitle={IJCAI},
  year={2020}
}

@inproceedings{bouton2020point,
  title={Point-Based Methods for Model Checking in Partially Observable Markov Decision Processes.},
  author={Bouton, Maxime and Tumova, Jana and Kochenderfer, Mykel J},
  booktitle={AAAI},
  year={2020}
}

@article{ahmadi2020stochastic,
  title={Stochastic finite state control of POMDPs with LTL specifications},
  author={Ahmadi, Mohamadreza and Sharan, Rangoli and Burdick, Joel W},
  journal={arXiv},
  year={2020}
}

@article{guo2015multi,
  title={Multi-agent plan reconfiguration under local LTL specifications},
  author={Guo, Meng and Dimarogonas, Dimos V},
  journal={IJRR},
  volume={34},
  number={2},
  year={2015},
  publisher={SAGE Publications Sage UK: London, England}
}

@inproceedings{svorenova2015temporal,
  title={Temporal logic motion planning using POMDPs with parity objectives: case study paper},
  author={Svore{\v{n}}ov{\'a}, M{\'a}ria and Chmel{\'\i}k, Martin and Leahy, Kevin and Eniser, Hasan Ferit and Chatterjee, Krishnendu and {\v{C}}ern{\'a}, Ivana and Belta, Calin},
  booktitle={HSCC},
  year={2015},
}

@inproceedings{guo2013revising,
  title={Revising Motion Planning Under Linear Temporal Logic Specifications in Partially Known Workspaces},
  author={Guo, Meng and Johansson, Karl H and Dimarogonas, Dimos V},
  booktitle={ICRA},
  year={2013}
}

@inproceedings{merlin2020locally,
  title={Locally Observable Markov Decision Processes},
  author={Merlin, Max and Parikh, Neev and Rosen, Eric and Konidaris, George},
  booktitle={ICRA 2020 Workshop on Perception, Action, Learning},
  year={2020}
}

@inproceedings{horak2019solving,
  title={Solving Partially Observable Stochastic Games with Public Observations},
  author={Hor{\'a}k, Karel and Bo{\v{s}}ansk{\`y}, Branislav},
  booktitle={AAAI},
  year={2019},
}

@article{madani1999pomdp,
  author = {Madani, Omid},
  journal={AAAI},
  year = {1999},
  title = {On the Computability of Infinite-Horizon Partially Observable Markov Decision Processes},
}

\end{document}